\documentclass[conference]{IEEEtran}
\usepackage{times}

\usepackage[numbers]{natbib}
\usepackage{multicol}
\usepackage{graphicx}
\usepackage{amsmath}
\usepackage{cuted}
\usepackage{float}     
\usepackage{afterpage} 
\usepackage{booktabs}
\usepackage{tabularx}
\usepackage{amssymb}
\usepackage{algorithm}
\usepackage{algorithmic}
\usepackage[bookmarks=true]{hyperref}
\usepackage{booktabs} 
\usepackage{tabularx} 
\usepackage{multirow} 
\usepackage{listings}
\usepackage{xcolor}

\definecolor{codegreen}{rgb}{0,0.6,0}
\definecolor{codegray}{rgb}{0.5,0.5,0.5}
\definecolor{codepurple}{rgb}{0.58,0,0.82}
\definecolor{backcolour}{rgb}{0.95,0.95,0.92}

\usepackage{caption}
\usepackage{arydshln}
\begin{document}

\title{\hspace{-1ex}MVP-Nav: Multi-layer Value Map Planner Navigator\hspace{-1ex}}


\author{
    \authorblockN{
        Wenyuan Xie, Shaokai Wu, Yijin Zhou, Yanbiao Ji, Guodong Zhang, \\
        Bayram Bayramli, Qiuchang Li, Xunchu Zhou, Yue Ding$^*$, and Hongtao Lu$^*$
    }
    \authorblockN{Shanghai Jiao Tong University}
    \authorblockN{$^*$Corresponding authors: dingyue@sjtu.edu.cn, htlu@sjtu.edu.cn}
}
\maketitle


\begin{abstract}
Zero-shot Object Goal Navigation (ZSON) with RGB-only perception poses a fundamental challenge for embodied agents, as the absence of explicit depth information introduces severe physical uncertainty and semantic–physical misalignment. Existing approaches either rely on high-level semantic reasoning without geometric grounding or learn end-to-end policies that lack explicit physical constraints, often resulting in semantically plausible but physically unsafe behaviors. In this paper, we propose \textbf{MVP-Nav}, \textbf{a physical-aware RGB-only navigation} framework that aligns perception, planning, and control with the real 3D world. \textbf{MVP-Nav} reconstructs explicit physical occupancy from monocular observations by leveraging 3D foundation models to project 2D semantic instances into 3D oriented bounding boxes, forming a global spatial semantic representation. To unify high-level semantic reasoning and low-level physical constraints, we introduce a Multi-layer Value Map (MVM) that integrates semantic priorities and reconstructed geometry into a shared cost space, enabling physically grounded geometric planning. Extensive experiments on zero-shot object navigation benchmarks demonstrate that \textbf{MVP-Nav} significantly outperforms existing depth-free methods, achieving state-of-the-art performance and validating that structured physical priors can effectively compensate for the absence of active depth sensors. 

\end{abstract}

\IEEEpeerreviewmaketitle

\section{Introduction}
\textbf{Zero-shot Object Goal Navigation} (ZSON) \citep{majumdar2022zson} is a fundamental capability for autonomous agents, requiring them to locate unseen objects in new environments based on natural language instructions. While conventional methods relied heavily on active sensors like LiDAR or RGB-D cameras to obtain precise metric information \citep{yin2024sg}, recent research has pivoted towards \textbf{RGB-only navigation} \citep{jin2025panonavmaplesszeroshotobject, cai2024bridging}. This shift is motivated not only by the desire to reduce hardware costs and enhance the accessibility of consumer-grade robots but also by the goal of challenging agents with higher-level visual reasoning. As noted by \citet{chaplot2020learning}, biological organisms primarily “rely on passive vision and internal priors to perceive 3D structure and navigate”; this paradigm therefore compels robots to move without depth sensors, instead mimicking human-like perception patterns, inferring complex semantic meanings and latent geometric structures from monocular or panoramic visual cues combined with spatial commonsense.

\begin{figure}
    \centering
\includegraphics[width=1.0\linewidth]{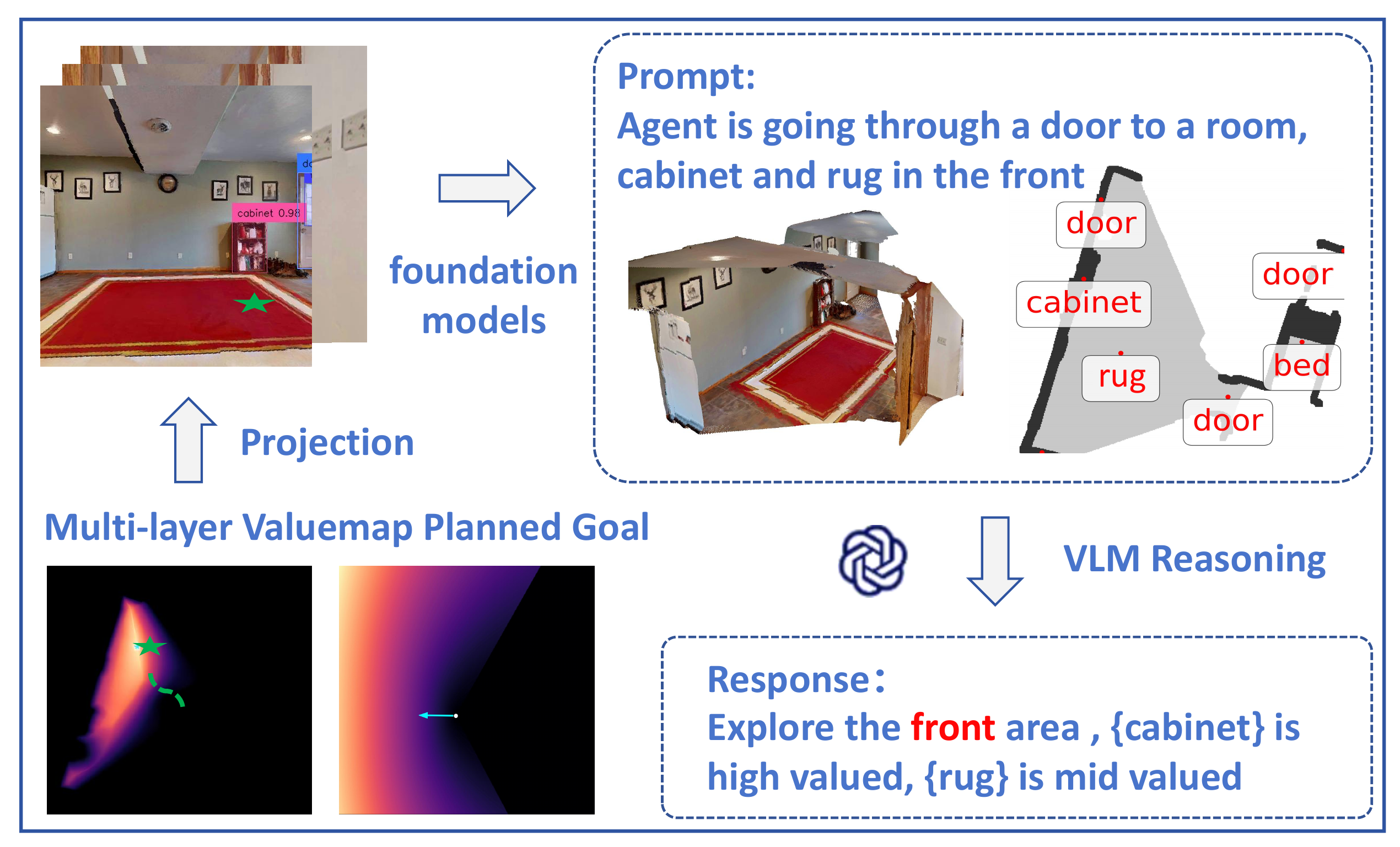}
    \caption{The overview of our navigation system.
    }
    \label{fig_tradeoff}
\end{figure}

  Current research in RGB-only navigation primarily follows two paradigms.
  Semantic-driven methods \citep{jin2025panonavmaplesszeroshotobject} utilize Multimodal Large Language Models (MLLMs) for panoramic parsing or heuristic exploration. However, treating space as a collection of discrete semantic labels often leads to distorted topological relationships under perspective changes. End-to-end methods \citep{cai2024bridging,xue2025omninav,hu2025astranav} learn spatial mappings through vision-action coupling.
  However, without explicit geometric modeling, they often fail to accurately perceive physical obstacles, resulting in navigation that is semantically reasonable yet physically unsafe.
  Collectively, these approaches expose a fundamental dilemma: existing solutions either decouple physical perception from high-level planning or sacrifice geometric constraints at the level of low-level control.
  As a result, current systems lack a unified representation that aligns discrete semantic reasoning with continuous physical boundaries.
  These limitations highlight a critical bottleneck in RGB-only navigation: the inability to recover explicit 3D scale from 2D sequences leads to \textbf{physical uncertainty} (i.e., the ambiguity of object distance and scale), and further induces \textbf{semantic-physical misalignment}, where high-level reasoning becomes disconnected from low-level physical constraints.

To address these challenges, we argue that the perception and control layers in object-goal navigation should align as closely as possible with the real physical world, while planning should leverage both historical exploration memory and real-time observations. 
Based on this principle, we propose \textbf{MVP-Nav (Multi-layer Value Map Planner Navigator)}. Our framework is motivated by the insight that physical occupancy in monocular vision can be reconstructed via geometric priors rather than explicit depth estimation. To mitigate \textbf{physical uncertainty}, we leverage 3D foundation models to project semantic instances from 2D observations into 3D space, representing them as Oriented Bounding Boxes (OBB) \citep{gottschalk2000collision}. This semantic-to-physical re-projection transforms ambiguous depth cues into deterministic spatial representations and enables the construction of a Global Spatial Semantic List (GSSL) to maintain the locations and scales of observed instances.
Furthermore, to address the \textbf{semantic-physical misalignment}, we introduce the Multi-layer Value Map (MVM) mechanism. MVM acts as a unified integration layer that embeds MLLM-generated high-level semantic priorities and low-level physical constraints into a shared cost space. This design enables geometric path planning (e.g., via the Fast Marching Method) that is jointly governed by reconstructed physical occupancy and semantic guidance. By decoupling high-level reasoning from low-level execution, \textbf{MVP-Nav} ensures navigation that is both goal-directed and physically grounded.

The contributions of this work are summarized as follows:
\begin{itemize}
    \item We propose a \textbf{physical-aware paradigm} for RGB-only ZSON, which reconstructs explicit 3D spatial occupancy from monocular observations by leveraging 3D foundation models and OBB-based fusion.
    \item We introduce the \textbf{Multi-layer Value Map (MVM)} architecture, providing a bridge to align discrete semantic reasoning with continuous physical constraints in a unified cost space.
    \item Extensive evaluations demonstrate that MVP-Nav achieves state-of-the-art performance among depth-free methods, proving that well-structured physical priors can effectively compensate for the absence of active depth sensors.
\end{itemize}

\section{Related works}

\subsection{Conventional Object Goal Navigation}
\textbf{Object Goal Navigation (ObjectNav)} requires an agent to locate and approach a specific object category in unseen environments. The development of this field is closely tied to the standardization of benchmarks, particularly the Habitat platform \cite{savva2019habitat}. By providing photorealistic 3D datasets such as Matterport3D and Gibson \cite{xiazamirhe2018gibsonenv}, Habitat established the ``Success weighted by Path Length'' (SPL) \cite{anderson2018evaluationembodiednavigationagents} as the primary metric to evaluate both navigation effectiveness and path efficiency. 

Historically, ObjectNav has been dominated by Geometry-driven Paradigms that prioritize explicit metric mapping and frontier exploration. Early successful agents primarily relied on active depth sensing (e.g., RGB-D) to build explicit 2D/3D occupancy representations for path planning \cite{chaplot2020object}. A classic representative is Stubborn \cite{9981646}, which effectively leverages geometric baselines to prioritize collision-free frontier exploration, demonstrating that structured physical occupancy is fundamental to reliable navigation. As the field progressed, the focus shifted from pure geometric obstacle avoidance to cognitive process modeling \cite{cao2025cognav}. This paradigm shift is deeply rooted in neuroscientific findings, which suggest that both humans and animals maintain internal representations of environmental geometry—such as grid cells and place cells—to support flexible spatial navigation \cite{zeng2022theory}. Consistent with these theories, recent studies have shown that human-like navigation involves maintaining and dynamically updating fine-grained cognitive states \cite{cao2025cognav}. This implies that while metric geometry provides a physical safety foundation, the integration of high-level environmental understanding is essential for efficient exploration. This evolution is further reflected in graph-based works like VoroNav \cite{wu2024voronav}, which utilizes Voronoi diagrams to abstract spatial topology while still leaning on the geometric scaffolding established by earlier depth-based methods. 

\subsection{Recent RGB-only Navigation}
The transition toward RGB-only Navigation represents a deliberate move toward simulating higher-level visual cognition, where agents must rely on semantic cues and spatial memory rather than direct distance measurements. Modern depth-free methods have evolved from early reactive policies into two main technical paradigms:

\paragraph*{Modular Reasoning based on Image/Pixel Space} 
Early RGB-only attempts like ONN \cite{yang2018visualsemanticnavigationusing}and Target-driven RL \cite{zhu2016targetdrivenvisualnavigationindoor} utilized visual scene priors or generic feature embeddings to guide exploration. To compensate for the lack of depth, modern works leverage Multi-modal Large Language Models (MLLMs) to perform reasoning directly on observed images or panoramic views \cite{cai2024bridging, jin2025panonavmaplesszeroshotobject, chen2026geometrically}. For instance, PixNav \cite{cai2024bridging} bridges foundation models and navigation via pixel-guided goal specification, while ImagineNav \cite{ICLR2025_eb261df4} and PanoNav \cite{jin2025panonavmaplesszeroshotobject} use 'scene imagination' and panoramic parsing to predict semantic waypoints. Similarly, SG-Nav \cite{yin2024sg} prompts LLMs with 3D scene graphs to enhance reasoning. A notable recent work, Mobility VLA \cite{pmlr-v270-xu25b}, leverages long-context reasoning with topological graph priors to navigate in known environments. However, these methods either depend on environment-specific priors or perform inference in 2D semantic spaces, frequently suffering from a Semantic-Physical Gap. In contrast, MVP-Nav targets zero-shot exploration in entirely unseen environments by reconstructing a 3D substrate for every decision, ensuring the agent's 'mental plan' is consistently grounded in the latent 3D layout. However, because these methods perform inference in 2D image semantics or abstract graph spaces, they frequently suffer from a Semantic-Physical Gap. The agent's 'mental plan' often fails to ground in the environment's latent 3D layout, leading to semantically logical but physically unreachable goal points.

\begin{figure*}
    \centering
\includegraphics[width=\linewidth]{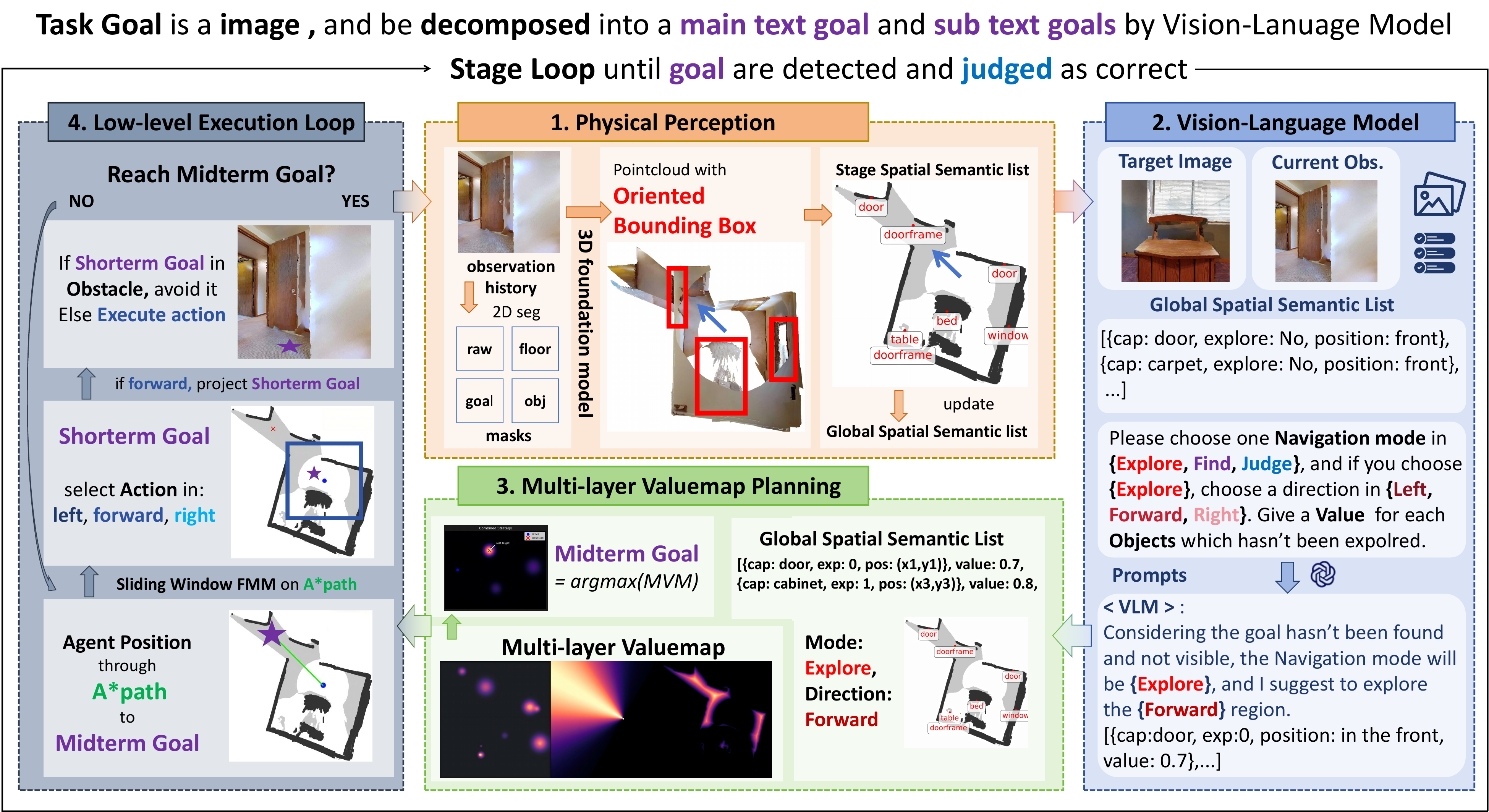}
    \caption{Overview of the \textbf{MVP-Nav} framework. Our system employs a recursive architecture that first transforms monocular RGB sequences into a 3D Global Spatial Semantic List (GSSL) using foundation models. A Vision-Language Model (VLM) then reasons over this list to assign semantic scores and determine the navigation mode. These insights are integrated into a Multi-layer Value Map (MVM) to identify an optimal midterm goal $g_{mid}$ within a unified cost space. Finally, a low-level execution loop performs $A^{*}$ and FMM planning, ensuring physical safety via real-time semantic floor re-projection.}
    \label{fig_method}
\end{figure*}

\paragraph*{End-to-End Learning Paradigm} 
This paradigm explores mapping raw RGB pixels directly to control signals, with early foundational works such as SAVN \cite{wortsman2019learninglearnlearnselfadaptive} focusing on self-adaptive navigation. Modern attempts like OmniNav \cite{xue2025omninav} and the navigation foundation model NavFoM \cite{zhang2025embodiednavigationfoundationmodel} seek to achieve generalizable navigation through large-scale pre-training across diverse tasks and embodiments. While these methods provide efficient, low-latency policies, they typically treat spatial relations as implicit latent features. Without an explicit persistent 3D substrate to ensure \textbf{Embodied Spatial Consistency}, these end-to-end models struggle to maintain stable environmental awareness over long horizons, manifesting the \textbf{Physical Uncertainty} and \textbf{Semantic-Physical Misalignment} that our work aims to address.

\vspace{5pt}
Unlike these paradigms, \textbf{MVP-Nav} reconciles high-level reasoning with physical boundaries by grounding VLM decisions within a recovered 3D volume via the Multi-layer Value Map. This transition from image-space imagination to physically-grounded planning ensures both goal-directed efficiency and environmental safety, fundamentally overcoming the bottleneck of depth-free navigation.

\section{Methodology}
\label{sec:methodology}

\subsection{Problem Statement}
\label{subsec:preliminaries}

Following the standard experimental protocols of Object Goal Navigation (ON) benchmarks \citep{savva2019habitat}, we consider an embodied agent operating within a large-scale, \textit{stationary} indoor environment $\mathcal{E}$. The agent is equipped with a reliable onboard localization system that provides its global pose estimate $p_t = (x_t, y_t, \theta_t)$ at each time step $t$. Despite the availability of pose information, the environmental geometry, semantic distribution, and the exact coordinates of the target remain entirely unknown at the start of the mission. 

The agent's perception is strictly constrained to monocular vision, where the observation tuple at time $t$ is $O_t = \{I_t, p_t\}$. We adhere to a rigorous \textit{RGB-only constraint}, meaning the system does not rely on any active depth sensors or LiDAR. Consequently, all spatial information must be inferred purely from the image sequence $\mathcal{I}$. The navigation goal is specified by a goal image $I_{goal}$ or a language instruction. The task is deemed successful if the agent executes a \texttt{STOP} command at a pose $p_t$ such that $\|p_t - p^*\|_2 < d$, where $p^*$ represents the target position and $d$ is a predefined distance threshold.

\subsection{Pipeline Overview}
\label{subsec:pipeline_overview}

In this subsection, we will introduce the overall process of the entire episode, including the start and end of tasks, as well as the multi round navigation mechanism in between. The navigation mechanism as the core algorithm will be explained in detail in the following content, while the start and end of tasks as non core content will only be introduced here.

\subsubsection{Episode Start}

The target image is parsed by a VLM into a combination of a main-target and several sub-targets, which will be utilized in subsequent navigation loop. At the start of the episode, the robot first performs a 360-degree in-place rotation to obtain the initial observation before proceeding to the actual navigation.

\subsubsection{MVP-Nav Loop}
The \textbf{MVP-Nav} framework adopts a decoupled, recursive architecture designed to bridge the gap between high-level semantic reasoning and low-level physical constraints. As illustrated in Fig.~\ref{fig_method} and Algorithm~\ref{navigation_loop}, the system alternates between high-level decision stages and low-level execution loops through four logical components:

\textbf{Physical Perception}. 
This module transforms the monocular image sequence $\mathcal{I}$ into a structured Global Spatial Semantic List (GSSL). We utilize 3D foundation models as the geometric backbone for end-to-end 3D reconstruction, recovering pseudo-depth and physical scale. 2D instances from open-vocabulary models are projected into 3D Oriented Bounding Boxes (OBB) to populate the GSSL.

\par \textbf{VLM-based High-level Reasoning}. 
The Vision-Language Model (VLM) acts as the system's cognitive core. Leveraging common-sense priors, the VLM processes GSSL entities and the navigation goal to evaluate spatial-semantic relevance. It assigns heuristic weights to entities and determines the navigation strategy (e.g., \textit{Explore}, \textit{Find}, or \textit{Judge}).

\par \textbf{Multi-layer Value Map (MVM) Planning}. 
This component unifies high-level VLM reasoning with low-level geometric constraints into a shared cost space. By projecting 3D OBBs and VLM-assigned weights onto a 2D grid, the system generates a multi-layer value map integrating semantic attraction, exploration guidance, and traversability. A path search on this map establishes a stable midterm goal ($g_{mid}$).

\par \textbf{Low-level Execution Loop}. 
Once $g_{mid}$ is established, the agent enters a high-frequency control loop, generating short-term goals ($g_{shorterm}$, simplified as $g_{st}$) via $A^*$ \citep{hart1968formal} and the Fast Marching Method (FMM) \citep{sethian1996fast}. To ensure safety under the \textit{RGB-only constraint}, we implement a Safety Verification step by back-projecting semantic floor masks to validate real-time traversability. The loop continues until $g_{mid}$ is reached, triggering the next reasoning cycle.

\subsubsection{Episode Finish}

As for the termination of the whole episode, if the reasoning stage give the mode as Judge, which means the agent seems find the goal. The robot performs a 360° panoramic scan. We utilize LightGlue to match the goal image with the most relevant historical views. Final task completion is only triggered after a Vision-Language Model (VLM) confirms the target's presence in the immediate vicinity. If a sub goal or main goal is misdetected, the system will give it a low score to stop agent from being attracted again.

\begin{algorithm}[H]
\caption{\textbf{MVP-Nav}: Recursive Navigation Loop}
\label{navigation_loop}
\begin{algorithmic}[1]
\STATE \textbf{Input:} $I_{goal}, \mathbf{p}_0$; \quad \textbf{Init:} $\mathcal{I}_{curr} \leftarrow$ \text{initial 360° scan};
\WHILE{mission not complete}
    \STATE Update GSSL via Physical Perception using $\mathcal{I}_{curr}$;
    \STATE Select NavMode and Direction through GSSL and $I_{goal}$;
    \STATE Synthesize Value Map, determine midterm goal $\mathbf{g}^*$;
    \STATE $\mathcal{I}_{next} \leftarrow \emptyset$;
    \WHILE{not reached $\mathbf{g}_{mid}$}
        \STATE Generate short-term goal $\mathbf{g}_{st}$ toward $\mathbf{g}_{mid}$;
        \STATE Execute Obstacle Avoidance using real-time observations and $\mathbf{g}_{st}$;
        \STATE Record $I_t$ into $\mathcal{I}_{next}$ and update $\mathbf{p}_t$;
    \ENDWHILE
    \STATE $\mathcal{I}_{curr} \leftarrow \mathcal{I}_{next}$;
\ENDWHILE
\end{algorithmic}

\end{algorithm}

\subsection{Physical Perception for 3D Reconstruction and GSSL}
\label{subsec:physical_perception}

\begin{figure*}
    \centering
\includegraphics[width=\linewidth]{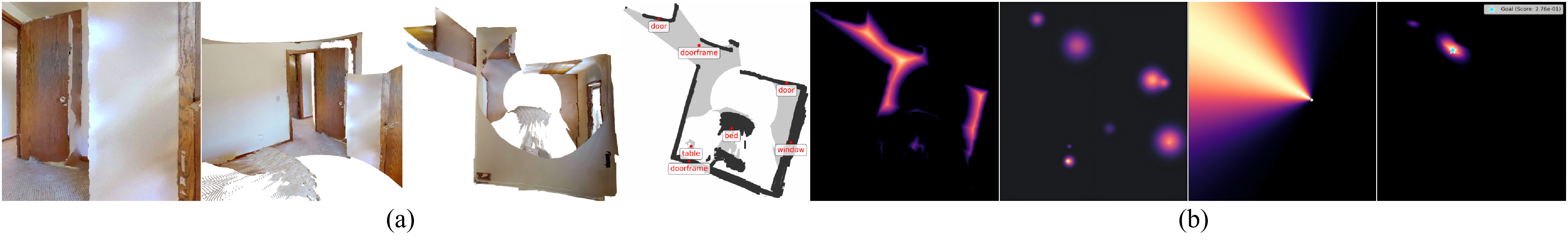}
    \caption{Construction of the Multi-layer Value Map (MVM). (a) The framework reconstructs 3D oriented bounding boxes (OBBs) from monocular RGB sequences to establish physical occupancy and spatial semantics. (b) These representations are integrated into the MVM, where semantic priorities and geometric constraints are fused into a unified cost space to determine the optimal navigation goal.}
    \label{mvm_long}
\end{figure*}

The Physical Perception module serves as the geometric and semantic foundation of \textbf{MVP-Nav}. As shown in Fig.~\ref{mvm_long} (a), this module aims to lift fleeting monocular RGB observations $I_t$ into a persistent 3D spatial memory through four stages.
It enables end-to-end 3D reconstruction from uncalibrated sequences by regressing globally consistent point clouds, thereby recovering pseudo-depth and physical scale from monocular observations.

For each navigation stage, the agent utilizes the visual geometry foundation model VGGT \citep{wang2025vggt} to process the image sequence $\mathcal{I}$, regressing the pseudo-depth map $\hat{z}$ and camera parameters $\hat{\mathbf{K}}$. All pixels $\mathbf{q} = [u, v]^\top$ from the original image $I_t$ are back-projected to construct a comprehensive local 3D point cloud $\mathcal{P}_{raw}$:
\begin{equation}
\mathbf{p} = \hat{z} \cdot \hat{\mathbf{K}}^{-1} [u, v, 1]^\top.
\end{equation}

In each stage, observation history collected in last stage are used as the source image of Physical Perception. To balance perceptual coverage and computational efficiency, we constrain the number of input images to approximately 30 per stage, if agent moves over 30 steps, this stage is stopped early. Empirically, exceeding 50 images significantly increases the risk of Out-of-Memory (OOM) errors during the global point cloud reconstruction phase.

Simultaneously, Grounded-SAM \citep{ren2024groundedsamassemblingopenworld,liu2023groundingdino,kirillov2023sam} generates three categories of semantic masks by inference 3 times separately , namely target objects ($M_{target}$), general objects ($M_{obj}$), and the floor ($M_{floor}$), which collectively guide the branching of point cloud data. Specifically, the raw point cloud $\mathcal{P}_{raw}$ is projected into the Bird's-Eye View (BEV) space, where non-floor occupancy is extracted and processed via morphological inflation to generate a local occupancy map for $A^*$ and FMM planning (Sec. \ref{subsec:execution}). Points corresponding to $M_{floor}$ are isolated and passed to the low-level execution loop as a benchmark for real-time semantic re-projection verification. To prevent semantic cross-contamination during spatial fusion, the system explicitly decouples point clusters covered by $M_{target}$ and $M_{obj}$ to fit separate Oriented Bounding Boxes (OBB). This isolation ensures that navigationally critical target entities remain distinct in the GSSL, as Fig.~\ref{fig:GSSL}, and are not erroneously merged with nearby general obstacles.

For GSSL entity generation and verification, the system implements a robust construction pipeline in the local space. The final entity label $L_{entity}$ is determined by a majority voting mechanism based on the mode of all associated detections. To ensure temporal consistency and filter out sensor noise, an observation frequency constraint is enforced, requiring each entity to be associated across at least three independent frames. Furthermore, OBBs serve as the primary geometric entries, effectively optimizing computational efficiency while maintaining high physical occupancy accuracy.

Building upon these locally consistent entities, the system must then integrate them into a globally metric-consistent framework. To this end, scale ambiguity is resolved by correlating the predicted trajectory $\Gamma_{vggt}$ and the metric trajectory $\Gamma_{loc}$. The similarity transform $\mathcal{T}(\mathbf{x}) = \sigma \mathbf{R} \mathbf{x} + \mathbf{t}$ is derived by minimizing the residual:
\begin{equation}
\min_{\sigma, \mathbf{R}, \mathbf{t}} \sum_{i=1}^{N} \| \mathbf{p}_{loc,i} - (\sigma \mathbf{R} \mathbf{p}_{vggt,i} + \mathbf{t}) \|^2.
\end{equation}

Once the scales are unified, the GSSL can be dynamically updated via 3D Intersection-over-Union (IoU). In this stage, if semantics are consistent and the spatial overlap exceeds a predefined threshold, a weighted fusion is triggered to refine the entity properties. The preliminary category decoupling mentioned above ensures that target entities remain semantically salient throughout this fusion process, ultimately providing a high-fidelity memory for VLM-based reasoning.


\begin{figure}[h]
\centering
\begin{minipage}{0.95\linewidth}
\begin{lstlisting}
[   {
    "id": "node_11",
    "caption": "doorframe",
    "position": "directly behind, 1.18m away",
    "explore": false,
    "exploration_score": 0.2,
    "cls_type": "node"
    },
    {
    "id": "node_12",
    "caption": "cabinet",
    "position": "to the back-left, 2.99m away",
    "explore": false,
    "exploration_score": 0.4,
    "cls_type": "node"
    },
    {
    "id": "sub_8",
    "caption": "shelf",
    "position": "directly behind, 3.04m away",
    "explore": false,
    "exploration_score": 0.8,
    "cls_type": "sub"
    },
    ...]
\end{lstlisting}
\end{minipage}
\caption{An example of a part of GSSL.}
\vspace{-10pt}
\label{fig:GSSL}
\end{figure}

\subsection{VLM-based High-level Reasoning}
\label{subsec:vlm_reasoning}

The VLM reasoning module serves as the cognitive core, taking the GSSL and navigation goal $\mathcal{G}$ as inputs to output semantic importance scores $\alpha_i$ for each entity and a navigation mode to guide spatial planning. To evaluate the exploration value, the Vision-Language Model (VLM, e.g., GPT-4o) leverages common-sense priors to assess the relevance of environmental entities to the goal. In order to optimize computational efficiency, the VLM performs scoring exclusively on newly detected entities within the GSSL, denoted as having an explored status of False. For each novel entity $\omega_i$, the VLM assigns an exploration value score $\alpha_i \in [0, 1]$ based on its likelihood of being the target or a significant semantic cue, which is formally defined as:
\begin{equation}
\alpha_i = \text{VLM}(\omega_i, \mathcal{G} \mid \text{Common-sense Priors}).
\end{equation}
These scores are stored in the GSSL and serve as weights for evaluating spatial attraction during the cost-map generation stage.

Beyond scoring, the VLM acts as a high-level arbitrator that dynamically determines the navigation mode based on the distribution of exploration values. When no high-value entities are identified in the GSSL, the system operates in an explore mode, where the VLM provides coarse directional suggestions such as left-front, front, or right-front. The system then enhances the exploration rewards of the corresponding regions in the cost map to guide the agent toward unmapped areas. Once the target or a strongly correlated cue, such as a pillow when searching for a bed, is detected, the system transitions to a find mode and plans a trajectory directly toward the spatial center of the entity. Finally, when the agent approaches the main target, it enters a judge mode to collect multi-view observations of the object. These visual features are compared with the goal image $I_{goal}$ by the VLM to verify whether the current entity fulfills the task requirements before terminating the search.

\subsection{Multi-layer Value Map Planning}
\label{subsec:mvm_planning}
As shown in Fig.~\ref{mvm_long} (b), the MVP module integrates the GSSL, VLM-derived scores $\alpha_i$, and the local occupancy map to synthesize a cost space. By synthesizing semantic, directional, and geometric constraints, the system generates a total value map $\Phi_{total}$ to determine the optimal midterm goal $g_{mid}$. Specifically, the planning module constructs three independent value layers in the 2D grid space to represent different planning constraints:

\begin{itemize}
    \item Semantic Value Map ($\Phi_{sem}$): This layer is generated by aggregating the spatial contributions of GSSL entities. To avoid over-saturation, the sum is clipped at 1.0:
    \begin{equation}
    \Phi_{sem}(\mathbf{p}) = \min \left( \sum_{i \in \text{GSSL}} \alpha_i \eta_i \cdot e^{ -\frac{\|\mathbf{p} - \mathbf{c}_i\|^2}{2\sigma_i^2} }, 1.0 \right),
    \end{equation}

    where $\mathbf{c}_i$ denotes the projected center, $\sigma_i$ is the standard deviation derived from the entity's physical extent, and $\eta_i$ represents whether the object $\mathbf{o}_i$ has been explored or not: $\eta_i$ = 1 for newly-detected, and $\eta_i$ = 0.5 for already detected.

    \item Directional Value Map ($\Phi_{dir}$): This layer biases movement toward the VLM-suggested orientation. Letting $\Delta\theta = \theta_{\mathbf{p}} - \theta_{target}$ represent the angular deviation, the map is formulated as:
    \begin{equation}
    \Phi_{dir}(\mathbf{p}) = \exp \left( -\frac{\Delta\theta^2}{2\sigma_{\theta}^2} \right) \cdot \mathbb{I}\left( |\Delta\theta| \le \frac{\pi}{2} \right),
    \end{equation}
    where $\sigma_\theta$ is the directional variance controlling the angular spread of semantic guidance, and $\mathbb{I}(\cdot)$ is an indicator function that restricts the exploration gain to the forward 180° semi-circular sector.

    \item Traversability Value Map ($\Phi_{trav}$): This layer is derived from the geometric occupancy map. We compute the distance $d_{min}(\mathbf{p})$ to the nearest obstacle for each cell to ensure safe navigation:
    \begin{equation}
    \Phi_{trav}(\mathbf{p}) = 1 - \exp(-k \cdot d_{min}(\mathbf{p})).
    \end{equation}
    where $k$ is the distance penalty weight: $k=1.0$ for obstacles, and $k=0.5$ for unknown regions
\end{itemize}

The final total value map $\Phi_{total}$ is synthesized through an element-wise product of these constituent layers, such that $\Phi_{total} = \Phi_{sem} \odot \Phi_{dir} \odot \Phi_{trav}$. The optimal midterm goal $g_{mid}$ is then determined by identifying the global maximum within this fused space, calculated as $g_{mid} = \arg\max_{\mathbf{p}} \Phi_{total}(\mathbf{p})$. In this framework, $g_{mid}$ serves as the fixed reference point for the current navigation stage, guiding the agent's local execution until a new high-level decision or an environmental update triggers a re-evaluation.

Notably, this multiplicative fusion mechanism distinguishes MVM from traditional Artificial Potential Fields (APF)\citep{1225434}. While APF typically generates control gradients by summing attractive and repulsive forces at the execution level, MVM operates at the planning level by performing element-wise multiplication. This ensures that the selected $g_{mid}$ strictly satisfies all semantic and physical constraints simultaneously---any area marked as non-traversable in $\Phi_{trav}$ (value of 0) will result in a zero total value, effectively eliminating candidates that are semantically attractive but physically unreachable.

\subsection{Low-level Execution Loop}
\label{subsec:execution}
\begin{figure}[!t]
    \centering
\includegraphics[width=\linewidth]{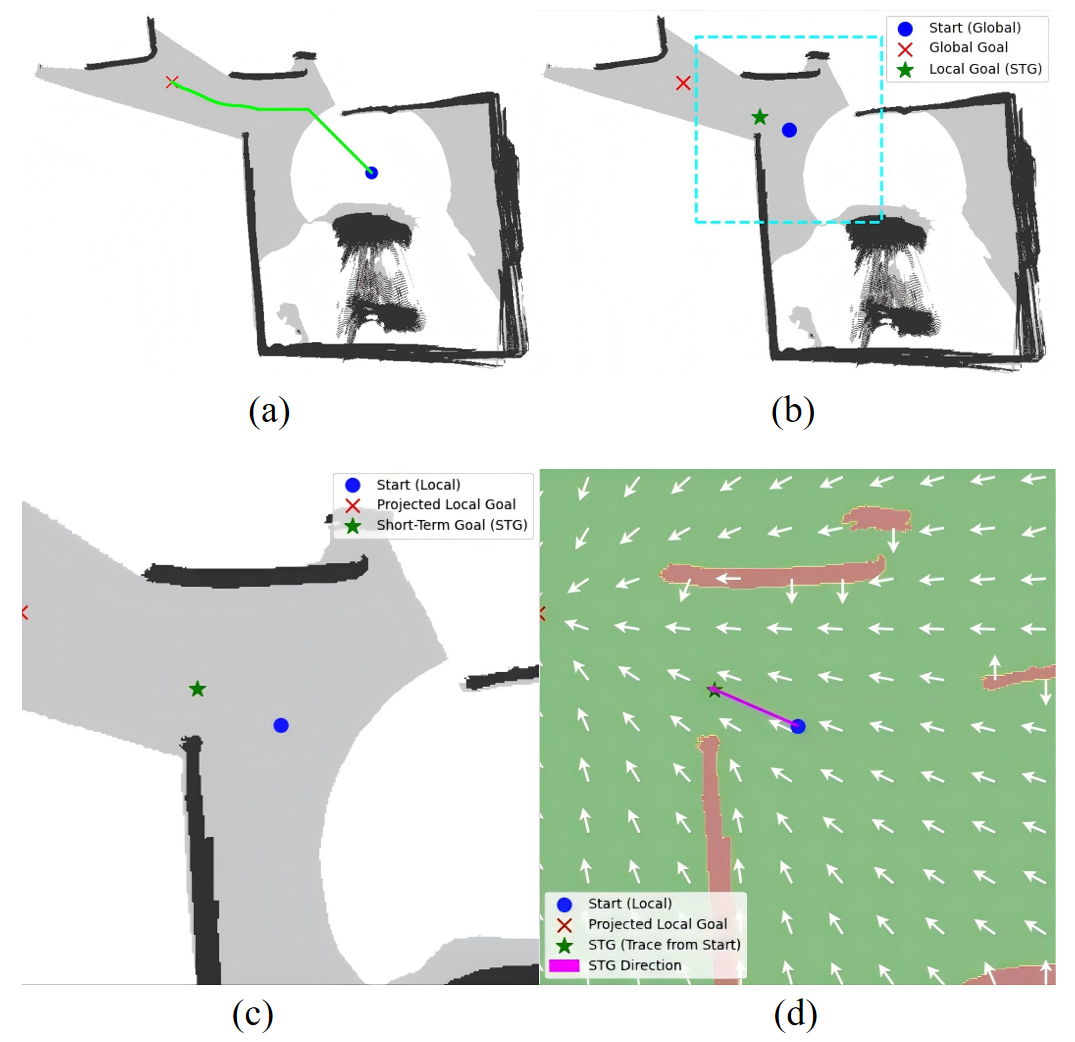}
    \caption{Visualization of the local goal $g_{st}$ generation. (a) The $A^*$ path from start position to the midterm goal. (b) The silding window of current position. (c) The intersection of sliding window and the $A^*$ path, which is the projected goal. (d) The System generates a short-term goal (STG) in the local occupancy map with FMM Planner.}
    \label{low_level}
\end{figure}

The execution layer translates $g_{mid}$ into motor commands by combining map-based pathfinding with reactive semantic verification. The system performs an $A^*$ search on the local occupancy map to establish a topological trajectory. To ensure high-frequency reactivity, as Fig.~\ref{low_level}, a sliding-window Fast Marching Method (FMM) identifies the intersection of the $A^*$ path and the window boundary as a projection target, generating a refined local goal $g_{st}$. While A* identifies the optimal discrete waypoint sequence, FMM solves the Eikonal equation on the MVM to generate a continuous potential field. This field acts as a safety-aware tracker, guiding the robot with smooth velocity commands while maintaining distance from identified hazards.

To compensate for map discretization, a safety layer uses the floor mask $M_{floor}$ for reactive avoidance. Before moving \texttt{FORWARD}, $g_{st}$ is re-projected onto the image plane:
\begin{equation}
\mathbf{q}_{local} = \hat{\mathbf{K}} \left( \mathbf{R} \cdot g_{st} + \mathbf{t} \right).
\end{equation}
If $\mathbf{q}_{local} \notin M_{floor}$, the agent halts and rotates until the goal aligns with the traversable region. This dual-layered strategy maintains global consistency alongside real-time constraints.

Furthermore, \textbf{MVP-Nav} employs a parallelized pipeline to eliminate transition latency. When $\|p_t - g_{mid}\| < \epsilon$, high-level reasoning for the next stage is triggered during active low-level control. Overlapping VLM inference with physical movement enables a fluid transition to $g_{mid, next}$ without requiring the agent to stop.

\begin{table*}[t]
\centering
\footnotesize
\resizebox{\textwidth}{!}{
\begin{tabular}{cccccccccc}
\toprule
\multirow{2.5}{*}{\textbf{Method}} & \multicolumn{3}{c}{\textbf{Navigation Setting}} & \multicolumn{2}{c}{\textbf{MP3D}} & \multicolumn{2}{c}{\textbf{HM3D \citep{ramakrishnan2021hm3d}}} & \multicolumn{2}{c}{\textbf{RoboTHOR}} \\
\cmidrule(lr){2-4} \cmidrule(lr){5-6} \cmidrule(lr){7-8} \cmidrule(lr){9-10}
& \textbf{Perception} & \textbf{Execution} & \textbf{Training-free} & SR (\%) & SPL (\%) & SR (\%) & SPL (\%) & SR (\%) & SPL (\%) \\
\midrule
SemExp \citep{chaplot2020object} & RGB-D & RGB-D & $\times$ & 36.0 & 14.4 & 37.9 & 18.8 & — & — \\
PONI \citep{ramakrishnan2022ponipotentialfunctionsobjectgoal} & RGB-D & RGB-D & $\times$ & 31.8 & 12.1 & — & — & — & — \\
Habitat-Web \citep{rramrakhya2022} & RGB-D & RGB-D & $\times$ & — & — & 41.5 & 16.0 & — & — \\
OVRL \citep{yadav2023offline} & RGB-D & RGB-D & $\times$ & — & — & 62.0 & 26.8 & — & — \\
\midrule
ESC \citep{zhou2023esc} & RGB-D & RGB-D & $\checkmark$ & 28.7 & 14.2 & 39.2 & 22.3 & 38.1 & 22.2 \\
VoroNav \citep{wu2024voronav} & RGB-D & RGB-D & $\checkmark$ & — & — & 42.0 & 26.0 & — & — \\
L3MVN \citep{10342512} & RGB-D & RGB-D & $\checkmark$ & 34.9 & 14.5 & 48.7 & 23.0 & 41.2 & 22.5 \\
OpenFMNav \citep{kuang2024openfmnav} & RGB-D & RGB-D & $\checkmark$ & 37.2 & 15.7 & 52.5 & 24.1 & 44.1 & 23.3 \\
VLFM \citep{yokoyama2024vlfm} & RGB-D & RGB-D & $\checkmark$ & 36.2 & 15.9 & 52.4 & 30.3 & 42.3 & 23.0 \\
SG-Nav \citep{yin2024sg} & RGB-D & RGB-D & $\checkmark$ & 40.2 & 16.0 & 54.0 & 24.9 & 47.5 & 24.0 \\
Unigoal \citep{yin2025unigoal} & RGB-D & RGB-D & $\checkmark$ & 41.6 & 16.4 & 54.5 & 25.1 & 48.0 & 24.2 \\
\midrule
ZSON \citep{majumdar2022zson} & RGB & RGB & $\times$ & 15.3 & 4.8 & 25.5 & 12.6 & — & — \\
PixNav \citep{cai2024bridging} & RGB & RGB & $\times$ & — & — & 37.9 & 20.5 & — & — \\
PanoNav \citep{jin2025panonavmaplesszeroshotobject} & Pano-RGB & Pano-RGB & $\checkmark$ & — & — & 43.5 & 23.7 & — & — \\
ImagineNav \citep{ICLR2025_eb261df4} & RGB & RGB-D & $\checkmark$ & — & — & 53.0 & 23.8 & — & — \\
\textbf{MVP-Nav (Ours)} & \textbf{RGB} & \textbf{RGB} & \textbf{$\checkmark$} & \textbf{50.4} & \textbf{18.1} & \textbf{65.4} & \textbf{27.9} & \textbf{57.5} & \textbf{26.8} \\
\bottomrule
\end{tabular}
}
\caption{Comprehensive comparison with state-of-the-art methods on Object-Goal Navigation benchmarks. We report Success Rate (SR) and SPL (\%) across three major datasets. Missing metrics are indicated by a dash (—).}
\label{benchmark}
\end{table*}

\begin{figure*}
    \centering
\includegraphics[width=\linewidth]{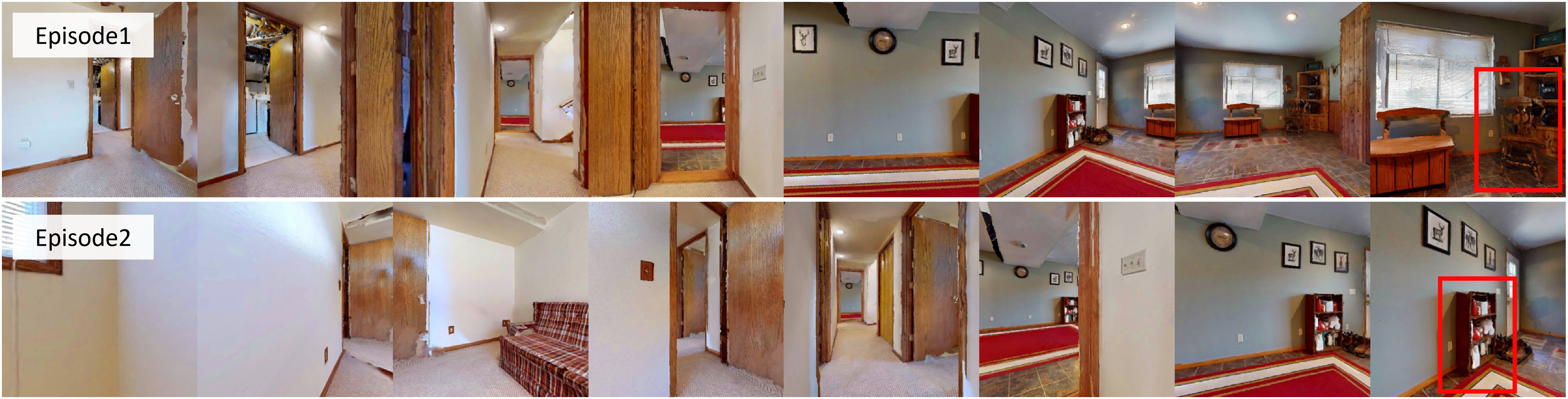}
    \caption{Examples of successful navigation episodes in the HM3D dataset. The trajectories illustrate how \textbf{MVP-Nav} effectively handles diverse indoor layouts.}
    \label{simulator_result}
\end{figure*}

\section{Experiments}
\label{sec:experiments}
\subsection{Experimental Setup}
\label{subsec:setup}

\paragraph*{Benchmarks} 
We evaluate the performance of \textbf{MVP-Nav} across three widely-adopted indoor object-goal navigation benchmarks, namely HM3D \citep{ramakrishnan2021hm3d}, MP3D \citep{Matterport3D}, and RoboTHOR \citep{deitke2020robothor}. Specifically, HM3D consists of 20 high-fidelity indoor scene reconstructions encompassing 2K validation episodes across 6 target object categories, while MP3D provides 11 real-world indoor environments with 1.8K validation episodes covering 20 categories. Additionally, we utilize RoboTHOR, which includes 1.8K validation episodes across 15 indoor environments involving 12 goal categories. In each episode, the goal image is defined by episode file in the dataset.

\paragraph*{Evaluation Metrics} 
To quantify navigation performance, we employ two standard metrics: Success Rate (SR) and Success weighted by Path Length (SPL). Specifically, SR represents the ratio of episodes where the agent successfully reaches the target object within a predefined distance threshold, while SPL evaluates navigation efficiency by normalizing the success rate with the ratio of the shortest path length to the actual distance traversed by the agent.

\paragraph*{Implementation Details} 
Experiments are conducted in the Habitat simulator with $640 \times 480$ RGB observations. The agent executes discrete actions ($0.25$m per step, $30^\circ$ rotations). We use VGGT for geometric perception and GPT-4o-mini for high-level reasoning. To assess deployment feasibility, all benchmarks and computational analyses are performed on a workstation with an NVIDIA RTX 6000 Ada GPU and an Intel Xeon Platinum 8352V CPU.

\subsection{Results}
\label{sec:quantitative}

 As summarized in Table~\ref{benchmark}, \textbf{MVP-Nav} demonstrates strong competitiveness across all benchmarks. Compared to the primary RGB-only baseline PanoNav, \textbf{MVP-Nav} exhibits an overwhelming performance gain; specifically, on the HM3D dataset, we significantly improve the Success Rate (SR) from 43.5\% to 65.4\% and the SPL from 23.7\% to 27.9\%. These results demonstrate that our multi-view geometric perception framework constructs spatial representations far more effectively than traditional monocular schemes. 

Notably, despite relying solely on RGB input, \textbf{MVP-Nav} achieves performance levels on par with, or even superior to, several state-of-the-art RGB-D systems equipped with physical depth sensors. For instance, our SR outperforms SG-Nav \citep{yin2024sg} (54.0\%) and UniGoal \citep{yin2025unigoal} (54.5\%), both of which utilize offline-trained scene graphs. This suggests that semantic planning based on physical constraints enables the system to reach a capability level close to that of active depth sensing. Fig.~\ref{simulator_result} shows two successful episodes in the HM3D dataset.

\subsection{Ablation Study and Analysis}

\paragraph*{Ablation Study}
As summarized in Table~\ref{ablation}, each core component in \textbf{MVP-Nav} is critical for balancing geometric fidelity and navigation intelligence.

\noindent \textbf{Ablation Analysis.} 
The failure of the ``Full-Sequence Reconstruction''  underscores our recursive framework's necessity. Maintaining global maps from long sequences causes severe cumulative drift, geometric warping, and Out-of-Memory (OOM) failures. In contrast, our recursive strategy reconstructs at smaller spatial scales, ensuring high-fidelity, drift-free local maps while preserving essential context via the Spatial Semantic List.

The semantic re-projection module is vital for operational safety. Removing it (w/o Sem. Re-proj.) marginally increases SPL ($27.9\% \rightarrow 30.5\%$) but significantly reduces Success Rate (SR). Without re-projection, agents adopt riskier, direct trajectories without verifying floor traversability; frequent collisions justify this efficiency trade-off for robust physical safety.

Furthermore, exploration mechanisms are critical for zero-shot navigation. Disabling exploration memory (w/o explore memory) causes SR to plummet to $38.2\%$ as agents become trapped in repetitive search. Similarly, removing LLM-based rating (w/o LLM rating) drops SR to $45.7\%$, demonstrating that without LLM-driven common-sense reasoning to prioritize candidate frontiers (e.g., identifying hallways as paths to kitchens), exploration becomes suboptimal, confirming the value of semantic anchoring.

\begin{table}[h]
\centering
\footnotesize
\setlength{\tabcolsep}{15pt} 
\begin{tabular}{lcc}
\toprule
\textbf{Module} & \textbf{SR (\%)} $\uparrow$ & \textbf{SPL (\%)} $\uparrow$ \\
\midrule
Full-Sequence Recon. & \multicolumn{2}{c}{OOM (Out of Memory)} \\
w/o Sem. Re-proj. & 53.2 & \textbf{30.5} \\
w/o explore memory & 38.2 & 13.4 \\
w/o LLM rating & 45.7 & 20.1 \\
\midrule
\textbf{full MVP-Nav} & \textbf{65.4} & 27.9 \\
\bottomrule
\end{tabular}
\caption{Ablation Results of \textbf{MVP-Nav} on the HM3D Dataset.}
\label{ablation}
\end{table}

\paragraph*{Component Selection Comparation}
We screened various foundation models to determine the optimal configuration for perception and reasoning, as detailed in Table~\ref{component}. Regarding 3D perception, while \textit{Depth-Anything V3} achieves a minimum latency of 2.56 s, its insufficient accuracy fails to provide reliable geometric constraints, leading to a 7.6\% drop in SR. Conversely, although \textit{Map-Anything} can generate dense maps, its VRAM consumption (28.78 GB) is prohibitive for real-time applications. The VGGT scheme provides the best balance, maintaining the highest SR with moderate resource usage. For the reasoning backend, GPT-4o-mini stands out with the lowest end-to-end latency (5.59 s) and superior logical robustness in following spatial semantic instructions compared to local alternatives like \textit{Llama3.2-vision}.

\begin{table}[htbp]
\centering
\footnotesize 
\begin{tabular*}{\linewidth}{@{\extracolsep{\fill}} lcccc}
\toprule
\quad \textbf{Module} & \textbf{SR (\%)} & \textbf{SPL (\%)} & \textbf{Latency} & \textbf{VRAM} \\
\midrule
\textit{Physical Perception} \\
\quad Depth-Anything V3 \citep{depthanything3} & 57.8 & 21.5 & \textbf{2.56} & 18.61 \\
\quad Map-Anything \citep{keetha2026mapanything} & 62.4 & 23.7 & 14.83 & 28.78 \\
\quad \textbf{VGGT} \citep{wang2025vggt} & \textbf{65.4} & \textbf{27.9} & 5.31 & \textbf{15.72} \\ 
\midrule
\textit{VLM Reasoning} \\
\quad Llama3.2-vision (Local) \citep{dubey2024llama} & 58.3 & 22.4 & \textbf{4.33} & -- \\
\quad Gemini-3-flash (API) \citep{team2023gemini} & 63.8 & 26.5 & 8.64 & -- \\
\quad \textbf{GPT-4o-mini (API)} \citep{hurst2024gpt} & \textbf{65.4} & \textbf{27.9} & 5.59 & -- \\ 
\bottomrule
\end{tabular*}
\caption{Performance analysis of each modules in \textbf{MVP-Nav}.}
\label{component}
\end{table}

\paragraph*{Computational Efficiency and Latency Analysis}
We evaluate the computational overhead by measuring the end-to-end latency of a single navigation loop. As reported in Table~\ref{latency}, the Physical Perception and VLM Reasoning module serves as the primary bottleneck, requiring 10.90 s per stage. 

To prevent this latency from causing motion interruptions, \textbf{MVP-Nav} utilizes a parallel pipeline mechanism that overlaps high-level reasoning with physical execution. Specifically, we initiate the planning for the next stage when the agent is within a Euclidean distance threshold of $\epsilon = 3$\,m from the current midterm goal. 

In each stage, the average step count remaining after reaching the distance $d = \epsilon$ is $10.3$. Given that the robot moves $0.25$\,m per step and the inference time is approximately $1.67$\,s, the buffer time of $10.3 \times 1.67\,\text{s} = 17.20\,\text{s}$ theoretically eliminates latency. In practice, however, the number of steps varies per stage, resulting in an actual average delay of $1.6$\,s. This is negligible compared to the average stage duration of $50.31$\,s. This minimal overhead demonstrates that our hierarchical decoupling successfully reconciles the high computational demands of MLLMs with the requirements for fluid indoor navigation.

\begin{table}[h]
\centering
\renewcommand{\arraystretch}{1.3} 
\begin{tabular}{cccc} 
\hline
\textbf{Module} & \textbf{Device} & \textbf{Latency (s)} & \textbf{Ratio (\%)} \\ \hline
Physical Perception & GPU & 5.31 & 40.26\% \\
MLLM Semantic Reasoning & API-Call & 5.59 & 42.38\% \\
Spatial Semantic List Update & CPU & 2.12 & 16.18\% \\
MVM \& Path Planning & CPU & 0.17 & 1.29\% \\ \hline
\textbf{Total} & - & \textbf{13.19} & \textbf{100\%} \\
Low-level Execution & \multicolumn{3}{c}{costs 1.67s per step}\\
\hline
\end{tabular}
\caption{Latency of different modules of \textbf{MVP-Nav}.}
\label{latency}
\end{table}

\vspace{-10pt}

\begin{figure*}
    \centering
\includegraphics[width=\linewidth]{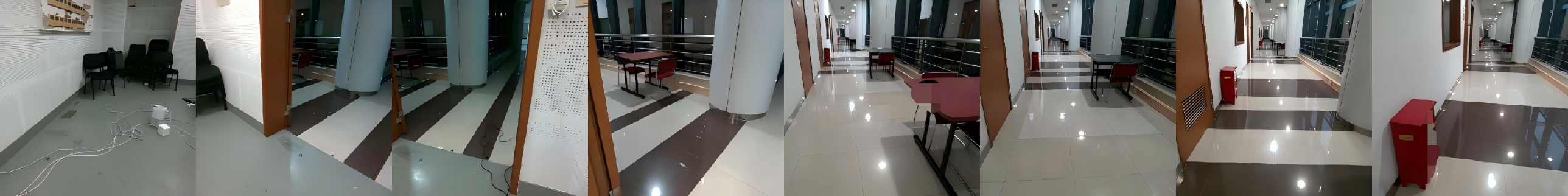}
    \caption{Real-world experimental results. We evaluated \textbf{MVP-Nav} on an Agibot G1 \citep{bu2025agibot} wheeled robot in a library corridor. The results (right) show that the system can successfully navigate to target objects over distances exceeding 15 meters, proving that our framework effectively translates from simulation to physical hardware.}
    \label{real_obs}
\end{figure*}

\subsection{Real-World Experiments}
\label{sec:realworld}

To bridge the gap between simulation and reality, we deploy \textbf{MVP-Nav} on a physical robotic platform to evaluate its navigational robustness in unstructured indoor environments.

\paragraph*{Hardware and Sensor Configuration}
We utilize the \textbf{Agibot G1}, a wheeled robot, for our real-world deployment. The hardware and sensor configurations are detailed as follows:
\begin{itemize}
    \item \textbf{Vision System}: Navigation is performed using only the RGB stream from the single head-mounted camera. To balance the field of view (FOV) between low-profile floor obstacles and eye-level semantic instances, the robot's head is fixed at a pitch angle of $-20^{\circ}$. 
    \item \textbf{Motion Tracking}: The G1 is equipped with high-precision wheel encoders and a 9-axis IMU. The onboard odometry system provides real-time pose feedback at a frequency of $1$\,kHz, which is critical for maintaining trajectory consistency during our recursive map updates.
    \item \textbf{Computing Infrastructure}: All heavy computations are offloaded to a dedicated server via a high-speed wireless link. The server is equipped with dual NVIDIA RTX 2080\,Ti GPUs.
\end{itemize}

\paragraph*{Experimental Environment and Setup}
The real-world evaluation is conducted in a university library corridor, a demanding environment characterized by long-range vistas, repetitive visual patterns, and varying lighting conditions. Specifically, we define two navigation tasks: (1) locating a \textbf{fire extinguisher box} situated in the left corridor after exiting the starting room, and (2) reaching a \textbf{large green plant} positioned along the wall of the right corridor. Each task was executed for 20 trials. These paths exceed $15$\,meters in length, with the left corridor stretching approximately $60$\,meters. To ensure stability, the robot's height was set to $130$\,cm. Another experiment scenario is an office with its many complex objects and its small range outdoor environment.In this scenario, the goal is a \textbf{computer on the chair} and and \textbf{trash bin} in front of the elevator

\begin{figure}[!t]
    \centering
\includegraphics[width=\linewidth]{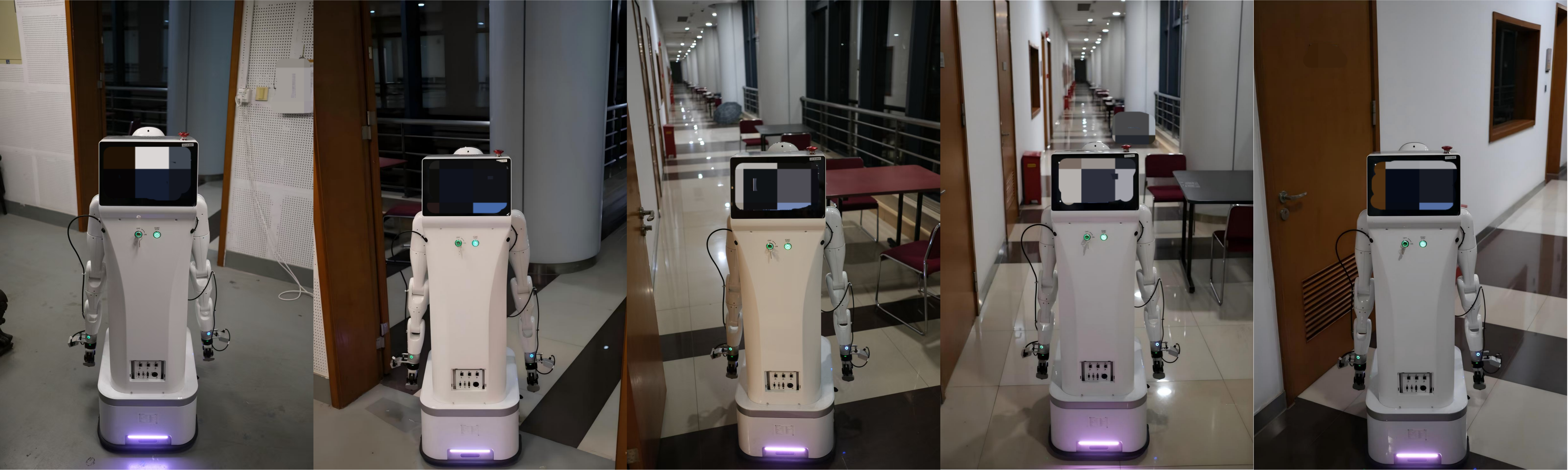}
    \caption{Real-world deployment on a wheeled robot platform. We present sequential frames captured from a follow-cam perspective to visualize the robot's physical execution. }
    \label{real_pic}
\includegraphics[width=\linewidth]{figures/fig_real_exp.pdf}
    \caption{Real-world deployment in another scenario.}
    \label{real_pic}
\end{figure}

\begin{table}[!t]
\centering
\renewcommand{\arraystretch}{1.3} 
\begin{tabular}{cccc} 
\hline
\textbf{Target} & \textbf{test times} & \textbf{success} & \textbf{Ratio (\%)} \\ \hline
Fire extinguisher box & 20 & 7 & 35.0 \\
Large green plant & 20 & 11 & 55.0 \\
\hline
\textbf{Total} & 40 & 18 & \textbf{45.0} \\ \hline
Computer on the chair & 20 & 12 & 60.0 \\
Trash bin & 20 & 7 & 35.0 \\
\hline
\textbf{Total} & 40 & 19 & \textbf{47.5} \\ \hline
\end{tabular}
\caption{Test results in two different real scenarios.}
\label{real_test}
\end{table}

\paragraph*{Results and Observations}
Despite the total absence of active depth sensors, \textbf{MVP-Nav} works successfully in the physical world, achieving an overall success rate of $45.0\%$ ($18/40$). In the right corridor task, the robot reached the green plant with a $55.0\%$ success rate ($11/20$). Performance in the left corridor was lower, with $7$ successful trials out of $20$ ($35.0\%$). Analysis suggests that the primary failure mode in the $60$-meter left corridor is the inherent difficulty of \textbf{long-range monocular depth estimation}, which occasionally leads to point cloud drifting and map distortion at extreme distances.

It is worth noting that while \textbf{MVP-Nav} was originally optimized for indoor benchmarks, these experiments were conducted in a challenging \textbf{semi-outdoor} setting due to library environment constraints. The fact that the system maintains a reasonable success rate under such conditions—where depth ambiguity is maximized—underscores its robustness. Our observations highlight two key strengths:

\begin{itemize}
    \item \textbf{Drift-Resistant Execution}: By leveraging the recursive stage-wise update mechanism, the cumulative drift from monocular vision input is effectively reduced. The robot maintains precise alignment between its $A^*$ path and the physical corridor boundaries.
    \item \textbf{Semantic Safety under Monocular Failure}: In the presence of glass partitions and textureless walls where monocular depth estimation frequently fails, \textbf{MVP-Nav} prevents collisions by identifying the semantic floor mask. The re-projection module successfully triggers proactive recovery whenever the projected goal falls outside the traversable floor region.
\end{itemize}

\section{Conclusion}
In this paper, we presented \textbf{MVP-Nav}, a training-free and RGB-only navigation framework that bridges the gap between high-level semantic reasoning and low-level physical occupancy. To address the challenge of achieving safe and consistent navigation without active depth sensing, we developed a 3D-driven OBB generation and computation algorithm that extracts structured geometric priors directly from 3D foundation models. These reconstructed OBBs are unified with MLLM-based semantic reasoning into a Multi-layer Value Map (MVM), enabling a tri-objective goal selection policy that respects both semantic relevance and physical constraints. Extensive experiments across multiple benchmarks and real-world deployment on a wheeled robot platform demonstrate that \textbf{MVP-Nav} generalizes robustly across diverse environments and instructions.
We also identify several technical insights for future improvement, including fine-tuning 3D foundation models with navigation-specific priors to further enhance reconstruction fidelity, and designing hybrid memory structures to balance implicit neural features with explicit and explainable geometric outputs. Furthermore, exploring a tighter coupling between physical features and the execution layer remains a key direction to further minimize latency and enhance the agility of sensor-minimalist embodied systems.

\bibliographystyle{plainnat}
\bibliography{references}

@inproceedings{savva2019habitat,
  title = {Habitat: A Platform for Embodied AI Research},
  author = {Savva, Manolis and Kadian, Abhishek and Maksymets, Oleksandr and Zhao, Yili and Wijmans, Erik and Jain, Bhavana and Straub, Julian and Liu, Jia and Koltun, Vladlen and Malik, Jitendra and Parikh, Devi and Batra, Dhruv},
  booktitle = {ICCV},
  year = {2019}
}

@article{liu2023groundingdino,
  title = {Grounding DINO: Marrying DINO with Grounded Pre-Training for Open-Set Object Detection},
  author = {Liu, Shilong and Zeng, Zhaoyang and Ren, Tianhe and Li, Feng and Zhang, Hao and Yang, Jie and Jiang, Qing and Li, Chunyuan and Yang, Jianwei and Su, Hang and Zhu, Jun and Zhang, Lei},
  journal = {arXiv:2303.05499},
  year = {2023}
}

@article{kirillov2023sam,
  title = {Segment Anything},
  author = {Kirillov, Alexander and Mintun, Eric and Ravi, Nikhila and Mao, Hanzi and Rolland, Chloe and Gustafson, Laura and Xiao, Tete and Whitehead, Spencer and Berg, Alexander C. and Lo, Wan-Yen and Doll{\'a}r, Piotr and Girshick, Ross},
  journal = {arXiv:2304.02643},
  year = {2023}
}

@inproceedings{wang2025vggt,
  title = {VGGT: Visual Geometry Grounded Transformer},
  author = {Wang, Jianyuan and Chen, Minghao and Karaev, Nikita and Vedaldi, Andrea and Rupprecht, Christian and Novotny, David},
  booktitle = {CVPR},
  year = {2025}
}

@article{chaplot2020object,
  title={Object goal navigation using goal-oriented semantic exploration},
  author={Chaplot, Devendra Singh and Gandhi, Dhiraj Prakashchand and Gupta, Abhinav and Salakhutdinov, Russ R},
  journal={Advances in Neural Information Processing Systems},
  volume={33},
  pages={4247--4258},
  year={2020}
}

@article{majumdar2022zson,
  title={Zson: Zero-shot object-goal navigation using multimodal goal embeddings},
  author={Majumdar, Arjun and Aggarwal, Gunjan and Devnani, Bhavika and Hoffman, Judy and Batra, Dhruv},
  journal={Advances in Neural Information Processing Systems},
  volume={35},
  pages={32340--32352},
  year={2022}
}

@inproceedings{zhou2023esc,
  title={Esc: Exploration with soft commonsense constraints for zero-shot object navigation},
  author={Zhou, Kaiwen and Zheng, Kaizhi and Pryor, Connor and Shen, Yilin and Jin, Hongxia and Getoor, Lise and Wang, Xin Eric},
  booktitle={International Conference on Machine Learning},
  pages={42829--42842},
  year={2023},
  organization={PMLR}
}

@article{wu2024voronav,
  title={Voronav: Voronoi-based zero-shot object navigation with large language model},
  author={Wu, Pengying and Mu, Yao and Wu, Bingxian and Hou, Yi and Ma, Ji and Zhang, Shanghang and Liu, Chang},
  journal={arXiv preprint arXiv:2401.02695},
  year={2024}
}

@article{kuang2024openfmnav,
  title={OpenFMNav: Towards Open-Set Zero-Shot Object Navigation via Vision-Language Foundation Models},
  author={Kuang, Yuxuan and Lin, Hai and Jiang, Meng},
  journal={arXiv preprint arXiv:2402.10670},
  year={2024}
}

@inproceedings{ramakrishnan2021hm3d,
  title={Habitat-Matterport 3D Dataset ({HM}3D): 1000 Large-scale 3D Environments for Embodied {AI}},
  author={Santhosh Kumar Ramakrishnan and Aaron Gokaslan and Erik Wijmans and Oleksandr Maksymets and Alexander Clegg and John M Turner and Eric Undersander and Wojciech Galuba and Andrew Westbury and Angel X Chang and Manolis Savva and Yili Zhao and Dhruv Batra},
  booktitle={Thirty-fifth Conference on Neural Information Processing Systems Datasets and Benchmarks Track},
  year={2021},
  url={https://arxiv.org/abs/2109.08238}
}

@article{Matterport3D,
  title={{Matterport3D}: Learning from {RGB-D} Data in Indoor Environments},
  author={Chang, Angel and Dai, Angela and Funkhouser, Thomas and Halber, Maciej and Niessner, Matthias and Savva, Manolis and Song, Shuran and Zeng, Andy and Zhang, Yinda},
  journal={International Conference on 3D Vision (3DV)},
  year={2017}
}

@inproceedings{deitke2020robothor,
  title={Robothor: An open simulation-to-real embodied ai platform},
  author={Deitke, Matt and Han, Winson and Herrasti, Alvaro and Kembhavi, Aniruddha and Kolve, Eric and Mottaghi, Roozbeh and Salvador, Jordi and Schwenk, Dustin and VanderBilt, Eli and Wallingford, Matthew and others},
  booktitle={Proceedings of the IEEE/CVF conference on computer vision and pattern recognition},
  pages={3164--3174},
  year={2020}
}

@inproceedings{cao2025cognav,
  title={Cognav: Cognitive process modeling for object goal navigation with llms},
  author={Cao, Yihan and Zhang, Jiazhao and Yu, Zhinan and Liu, Shuzhen and Qin, Zheng and Zou, Qin and Du, Bo and Xu, Kai},
  booktitle={Proceedings of the IEEE/CVF International Conference on Computer Vision},
  pages={9550--9560},
  year={2025}
}

@article{yin2024sg,
  title={Sg-nav: Online 3d scene graph prompting for llm-based zero-shot object navigation},
  author={Yin, Hang and Xu, Xiuwei and Wu, Zhenyu and Zhou, Jie and Lu, Jiwen},
  journal={Advances in neural information processing systems},
  volume={37},
  pages={5285--5307},
  year={2024}
}

@inproceedings{yin2025unigoal,
  title={Unigoal: Towards universal zero-shot goal-oriented navigation},
  author={Yin, Hang and Xu, Xiuwei and Zhao, Linqing and Wang, Ziwei and Zhou, Jie and Lu, Jiwen},
  booktitle={Proceedings of the Computer Vision and Pattern Recognition Conference},
  pages={19057--19066},
  year={2025}
}

@misc{jin2025panonavmaplesszeroshotobject,
      title={PanoNav: Mapless Zero-Shot Object Navigation with Panoramic Scene Parsing and Dynamic Memory}, 
      author={Qunchao Jin and Yilin Wu and Changhao Chen},
      year={2025},
      eprint={2511.06840},
      archivePrefix={arXiv},
      primaryClass={cs.CV},
      url={https://arxiv.org/abs/2511.06840}, 
}

@inproceedings{cai2024bridging,
  title={Bridging zero-shot object navigation and foundation models through pixel-guided navigation skill},
  author={Cai, Wenzhe and Huang, Siyuan and Cheng, Guangran and Long, Yuxing and Gao, Peng and Sun, Changyin and Dong, Hao},
  booktitle={2024 IEEE International Conference on Robotics and Automation (ICRA)},
  pages={5228--5234},
  year={2024},
  organization={IEEE}
}

@inproceedings{ICLR2025_eb261df4,
 author = {Zhao, Xinxin and Cai, Wenzhe and Tang, Likun and Wang, Teng},
 booktitle = {International Conference on Learning Representations},
 editor = {Y. Yue and A. Garg and N. Peng and F. Sha and R. Yu},
 pages = {94387--94401},
 title = {ImagineNav: Prompting Vision-Language Models as Embodied Navigator through Scene Imagination},
 url = {https://proceedings.iclr.cc/paper_files/paper/2025/file/eb261df4322a8bd0a73093c4d8a0d02d-Paper-Conference.pdf},
 volume = {2025},
 year = {2025}
}

@inproceedings{chaplot2020learning,
  title={Learning To Explore Using Active Neural SLAM},
  author={Chaplot, Devendra Singh and Gandhi, Dhiraj and Gupta, Saurabh and Gupta, Abhinav and Salakhutdinov, Ruslan},
  booktitle={International Conference on Learning Representations (ICLR)},
  year={2020}
}

@article{hu2025astranav,
  title={AstraNav-World: World Model for Foresight Control and Consistency},
  author={Hu, Junjun and Chen, Jintao and Bai, Haochen and Luo, Minghua and Xie, Shichao and Chen, Ziyi and Liu, Fei and Chu, Zedong and Xue, Xinda and Ren, Botao and others},
  journal={arXiv preprint arXiv:2512.21714},
  year={2025}
}

@article{xue2025omninav,
  title={OmniNav: A Unified Framework for Prospective Exploration and Visual-Language Navigation},
  author={Xue, Xinda and Hu, Junjun and Luo, Minghua and Shichao, Xie and Chen, Jintao and Xie, Zixun and Kuichen, Quan and Wei, Guo and Xu, Mu and Chu, Zedong},
  journal={arXiv preprint arXiv:2509.25687},
  year={2025}
}

@article{depthanything3,
  title={Depth Anything 3: Recovering the visual space from any views},
  author={Haotong Lin and Sili Chen and Jun Hao Liew and Donny Y. Chen and Zhenyu Li and Guang Shi and Jiashi Feng and Bingyi Kang},
  journal={arXiv preprint arXiv:2511.10647},
  year={2025}
}

@inproceedings{keetha2026mapanything,
  title={{MapAnything}: Universal Feed-Forward Metric {3D} Reconstruction},
  author={Nikhil Keetha and Norman M\"{u}ller and Johannes Sch\"{o}nberger and Lorenzo Porzi and Yuchen Zhang and Tobias Fischer and Arno Knapitsch and Duncan Zauss and Ethan Weber and Nelson Antunes and Jonathon Luiten and Manuel Lopez-Antequera and Samuel Rota Bul\`{o} and Christian Richardt and Deva Ramanan and Sebastian Scherer and Peter Kontschieder},
  booktitle={International Conference on 3D Vision (3DV)},
  year={2026},
  organization={IEEE}
}

@misc{ren2024groundedsamassemblingopenworld,
      title={Grounded SAM: Assembling Open-World Models for Diverse Visual Tasks}, 
      author={Tianhe Ren and Shilong Liu and Ailing Zeng and Jing Lin and Kunchang Li and He Cao and Jiayu Chen and Xinyu Huang and Yukang Chen and Feng Yan and Zhaoyang Zeng and Hao Zhang and Feng Li and Jie Yang and Hongyang Li and Qing Jiang and Lei Zhang},
      year={2024},
      eprint={2401.14159},
      archivePrefix={arXiv},
      primaryClass={cs.CV},
      url={https://arxiv.org/abs/2401.14159}, 
}

@misc{ramakrishnan2022ponipotentialfunctionsobjectgoal,
      title={PONI: Potential Functions for ObjectGoal Navigation with Interaction-free Learning}, 
      author={Santhosh Kumar Ramakrishnan and Devendra Singh Chaplot and Ziad Al-Halah and Jitendra Malik and Kristen Grauman},
      year={2022},
      eprint={2201.10029},
      archivePrefix={arXiv},
      primaryClass={cs.CV},
      url={https://arxiv.org/abs/2201.10029}, 
}

@inproceedings{rramrakhya2022,
  title={Habitat-Web: Learning Embodied Object-Search Strategies
         from Human Demonstrations at Scale},
  author={Ram Ramrakhya and Eric Undersander and Dhruv Batra and Abhishek Das},
  year={2022},
  booktitle={CVPR},
}

@inproceedings{yadav2023offline,
  title={Offline visual representation learning for embodied navigation},
  author={Yadav, Karmesh and Ramrakhya, Ram and Majumdar, Arjun and Berges, Vincent-Pierre and Kuhar, Sachit and Batra, Dhruv and Baevski, Alexei and Maksymets, Oleksandr},
  booktitle={Workshop on Reincarnating Reinforcement Learning at ICLR 2023},
  year={2023}
}

@INPROCEEDINGS{10342512,
  author={Yu, Bangguo and Kasaei, Hamidreza and Cao, Ming},
  booktitle={2023 IEEE/RSJ International Conference on Intelligent Robots and Systems (IROS)}, 
  title={L3MVN: Leveraging Large Language Models for Visual Target Navigation}, 
  year={2023},
  volume={},
  number={},
  pages={3554-3560},
  doi={10.1109/IROS55552.2023.10342512}}

@inproceedings{yokoyama2024vlfm,
  title={VLFM: Vision-Language Frontier Maps for Zero-Shot Semantic Navigation},
  author={Naoki Yokoyama and Sehoon Ha and Dhruv Batra and Jiuguang Wang and Bernadette Bucher},
  booktitle={International Conference on Robotics and Automation (ICRA)},
  year={2024}
}

@INPROCEEDINGS{9981646,
  author={Luo, Haokuan and Yue, Albert and Hong, Zhang-Wei and Agrawal, Pulkit},
  booktitle={2022 IEEE/RSJ International Conference on Intelligent Robots and Systems (IROS)}, 
  title={Stubborn: A Strong Baseline for Indoor Object Navigation}, 
  year={2022},
  volume={},
  number={},
  pages={3287-3293},
  doi={10.1109/IROS47612.2022.9981646}}

@inproceedings{xiazamirhe2018gibsonenv,
  title={Gibson {Env}: real-world perception for embodied agents},
  author={Xia, Fei and R. Zamir, Amir and He, Zhi-Yang and Sax, Alexander and Malik, Jitendra and Savarese, Silvio},
  booktitle={Computer Vision and Pattern Recognition (CVPR), 2018 IEEE Conference on},
  year={2018},
  organization={IEEE}
}

@misc{anderson2018evaluationembodiednavigationagents,
      title={On Evaluation of Embodied Navigation Agents}, 
      author={Peter Anderson and Angel Chang and Devendra Singh Chaplot and Alexey Dosovitskiy and Saurabh Gupta and Vladlen Koltun and Jana Kosecka and Jitendra Malik and Roozbeh Mottaghi and Manolis Savva and Amir R. Zamir},
      year={2018},
      eprint={1807.06757},
      archivePrefix={arXiv},
      primaryClass={cs.AI},
      url={https://arxiv.org/abs/1807.06757}, 
}

@misc{zhang2025embodiednavigationfoundationmodel,
      title={Embodied Navigation Foundation Model}, 
      author={Jiazhao Zhang and Anqi Li and Yunpeng Qi and Minghan Li and Jiahang Liu and Shaoan Wang and Haoran Liu and Gengze Zhou and Yuze Wu and Xingxing Li and Yuxin Fan and Wenjun Li and Zhibo Chen and Fei Gao and Qi Wu and Zhizheng Zhang and He Wang},
      year={2025},
      eprint={2509.12129},
      archivePrefix={arXiv},
      primaryClass={cs.RO},
      url={https://arxiv.org/abs/2509.12129}, 
}

@misc{wortsman2019learninglearnlearnselfadaptive,
      title={Learning to Learn How to Learn: Self-Adaptive Visual Navigation Using Meta-Learning}, 
      author={Mitchell Wortsman and Kiana Ehsani and Mohammad Rastegari and Ali Farhadi and Roozbeh Mottaghi},
      year={2019},
      eprint={1812.00971},
      archivePrefix={arXiv},
      primaryClass={cs.CV},
      url={https://arxiv.org/abs/1812.00971}, 
}

@misc{yang2018visualsemanticnavigationusing,
      title={Visual Semantic Navigation using Scene Priors}, 
      author={Wei Yang and Xiaolong Wang and Ali Farhadi and Abhinav Gupta and Roozbeh Mottaghi},
      year={2018},
      eprint={1810.06543},
      archivePrefix={arXiv},
      primaryClass={cs.CV},
      url={https://arxiv.org/abs/1810.06543}, 
}

@misc{zhu2016targetdrivenvisualnavigationindoor,
      title={Target-driven Visual Navigation in Indoor Scenes using Deep Reinforcement Learning}, 
      author={Yuke Zhu and Roozbeh Mottaghi and Eric Kolve and Joseph J. Lim and Abhinav Gupta and Li Fei-Fei and Ali Farhadi},
      year={2016},
      eprint={1609.05143},
      archivePrefix={arXiv},
      primaryClass={cs.CV},
      url={https://arxiv.org/abs/1609.05143}, 
}

@article{zeng2022theory,
  title={A theory of geometry representations for spatial navigation},
  author={Zeng, Taiping and Si, Bailu and Feng, Jianfeng},
  journal={Progress in Neurobiology},
  volume={211},
  pages={102228},
  year={2022},
  publisher={Elsevier}
}

@book{gottschalk2000collision,
  title={Collision queries using oriented bounding boxes},
  author={Gottschalk, Stefan Aric},
  year={2000},
  publisher={The University of North Carolina at Chapel Hill}
}

@article{hart1968formal,
  title={A formal basis for the heuristic determination of minimum cost paths},
  author={Hart, Peter E and Nilsson, Nils J and Raphael, Bertram},
  journal={IEEE transactions on Systems Science and Cybernetics},
  volume={4},
  number={2},
  pages={100--107},
  year={1968},
  publisher={IEEE}
}

@article{sethian1996fast,
  title={A fast marching level set method for monotonically advancing fronts.},
  author={Sethian, James A},
  journal={proceedings of the National Academy of Sciences},
  volume={93},
  number={4},
  pages={1591--1595},
  year={1996}
}

@article{dubey2024llama,
  title={The llama 3 herd of models},
  author={Dubey, Abhimanyu and Jauhri, Abhinav and Pandey, Abhinav and Kadian, Abhishek and Al-Dahle, Ahmad and Letman, Aiesha and Mathur, Akhil and Schelten, Alan and Yang, Amy and Fan, Angela and others},
  journal={arXiv e-prints},
  pages={arXiv--2407},
  year={2024}
}

@article{team2023gemini,
  title={Gemini: a family of highly capable multimodal models},
  author={Team, Gemini and Anil, Rohan and Borgeaud, Sebastian and Alayrac, Jean-Baptiste and Yu, Jiahui and Soricut, Radu and Schalkwyk, Johan and Dai, Andrew M and Hauth, Anja and Millican, Katie and others},
  journal={arXiv preprint arXiv:2312.11805},
  year={2023}
}

@article{hurst2024gpt,
  title={Gpt-4o system card},
  author={Hurst, Aaron and Lerer, Adam and Goucher, Adam P and Perelman, Adam and Ramesh, Aditya and Clark, Aidan and Ostrow, AJ and Welihinda, Akila and Hayes, Alan and Radford, Alec and others},
  journal={arXiv preprint arXiv:2410.21276},
  year={2024}
}

@article{bu2025agibot,
  title={Agibot world colosseo: A large-scale manipulation platform for scalable and intelligent embodied systems},
  author={Bu, Qingwen and Cai, Jisong and Chen, Li and Cui, Xiuqi and Ding, Yan and Feng, Siyuan and Gao, Shenyuan and He, Xindong and Hu, Xuan and Huang, Xu and others},
  journal={arXiv preprint arXiv:2503.06669},
  year={2025}
}

@InProceedings{pmlr-v270-xu25b,
  title = 	 {Mobility VLA: Multimodal Instruction Navigation with Long-Context VLMs and Topological Graphs},
  author =       {Xu, Zhuo and Chiang, Hao-Tien Lewis and Fu, Zipeng and Jacob, Mithun George and Zhang, Tingnan and Lee, Tsang-Wei Edward and Yu, Wenhao and Schenck, Connor and Rendleman, David and Shah, Dhruv and Xia, Fei and Hsu, Jasmine and Hoech, Jonathan and Florence, Pete and Kirmani, Sean and Singh, Sumeet and Sindhwani, Vikas and Parada, Carolina and Finn, Chelsea and Xu, Peng and Levine, Sergey and Tan, Jie},
  booktitle = 	 {Proceedings of The 8th Conference on Robot Learning},
  pages = 	 {3866--3887},
  year = 	 {2025},
  editor = 	 {Agrawal, Pulkit and Kroemer, Oliver and Burgard, Wolfram},
  volume = 	 {270},
  series = 	 {Proceedings of Machine Learning Research},
  month = 	 {06--09 Nov},
  publisher =    {PMLR},
  url = 	 {https://proceedings.mlr.press/v270/xu25b.html}
}

@INPROCEEDINGS{1225434,
  author={Min Cheol Lee and Min Gyu Park},
  booktitle={Proceedings 2003 IEEE/ASME International Conference on Advanced Intelligent Mechatronics (AIM 2003)}, 
  title={Artificial potential field based path planning for mobile robots using a virtual obstacle concept}, 
  year={2003},
  volume={2},
  number={},
  pages={735-740 vol.2},
  doi={10.1109/AIM.2003.1225434}}

@inproceedings{chen2026geometrically,
  title={Geometrically-constrained agent for spatial reasoning},
  author={Chen, Zeren and Lu, Xiaoya and Zheng, Zhijie and Li, Pengrui and He, Lehan and Zhou, Yijin and Shao, Jing and Zhuang, Bohan and Sheng, Lu},
  booktitle={Proceedings of the IEEE/CVF Conference on Computer Vision and Pattern Recognition},
  pages={38689--38699},
  year={2026}
}

\end{document}